\pdfoutput=1

\documentclass[11pt]{article}

\usepackage[]{naacl2021}

\usepackage{times}
\usepackage{latexsym}

\usepackage[T1]{fontenc}

\usepackage[utf8]{inputenc}

\usepackage{microtype}

\usepackage{soul}
\usepackage{amsmath}
\usepackage{graphicx}
\usepackage{booktabs}
\usepackage{caption}
\usepackage{subcaption}
\usepackage{xcolor}
\usepackage{enumitem,kantlipsum}

\title{Estimating Redundancy in Clinical Text}

\author{Thomas Searle$^1$, Zina Ibrahim$^1$, James Teo$^2$, Richard JB Dobson$^{1,3}$ \\
  $^1$Department of Biostatistics and Health Informatics, Institute of Psychiatry, \\ Psychology and Neuroscience, King’s College London, London, U.K.\\
  $^2$King's College Hospital NHS Foundation Trust, London, UK \\
  $^3$Institute of Health Informatics, University College London,\\
  London, London, U.K.\\
    \{firstname.lastname\}@kcl.ac.uk}

\begin{document}
\maketitle
\begin{abstract}
The current mode of use of Electronic Health Records (EHR) elicits text redundancy. Clinicians often populate new documents by duplicating existing notes, then updating accordingly. Data duplication can lead to propagation of errors, inconsistencies and misreporting of care. Therefore, measures to quantify information redundancy play an essential role in evaluating innovations that operate on clinical narratives. 

This work is a quantitative examination of information redundancy in EHR notes. We present and evaluate two methods to measure redundancy: an information-theoretic approach and a lexicosyntactic and semantic model. Our first measure trains large Transformer-based language models using clinical text from a large openly available US-based ICU dataset and a large multi-site UK based Hospital. By comparing the  information-theoretic \textit{efficient encoding} of clinical text against open-domain corpora, we find that clinical text is $\sim1.5$x to $\sim3$x less efficient than open-domain corpora at conveying \textit{information}. Our second measure, evaluates automated summarisation metrics Rouge and BERTScore to evaluate successive note pairs demonstrating lexicosyntactic and semantic redundancy, with averages from $\sim$43 to $\sim$65\%.

\end{abstract}

\section{Introduction}
Electronic Health Record (EHR) text details patient history, findings, symptoms, diagnoses, procedures and plans for future care. A single inpatient hospital stay can result in multiple document types (e.g. GP letters, inpatient admission / discharge notes) created by the different specialisms involved in the patient's care (e.g. nursing, A\&E, cardiology, neurology, radiology etc.) as well as progress documents to address previous questions and introducing follow-up actions or queries. As a result, a patient's records can contain different perspectives accumulated through time, by various specialities documenting the patient's `progress' throughout the care pathway \cite{Mathioudakis2016-dn}. Therefore, it naturally follows that EHR text and the design of systems induces redundancy. This is not necessarily a negative as repeated mentions could be used to indicate importance, corroboration or confirmation of a prior finding, diagnosis etc. However, using the clinical narratives for direct patient care can be difficult \cite{Kroth2018-uy}, as clinicians must navigate through potentially redundant, out-of-date or erroneous information to come to the \textit{current} state of a patient, although this problem of navigation and data consumption is not exclusive to unstructured portion of EHRs. For secondary research purposes \cite{Bayley2013-ek,Miriovsky2012-ut} this requires significant time cleaning and pre-processing data \cite{Miotto2016-wy,Landi2020-ne}. 

Using clinical narratives in EHRs is unavoidable. For direct patient care, forcing EHR users to specify patient state in only structured fields thereby avoiding free-text input is both impractical and insufficient  \cite{Goossen2011-yy,Abernethy2017-rz} and also does not consider existing free-text patient data. Outside of direct patient care, prior work has shown EHR text analysis offers insights in diverse areas such as disease classification \cite{Perlis2012-dx}, trajectory modelling \cite{Paik2019-pv}, patient stratification \cite{Landi2020-ne}, therapeutic development \cite{Maudsley2018-yv} and personalised medicine \cite{Topol2019-ja}. Yet, the free-text content of EHRs forces researchers to spend considerable time manually exploring datasets attempting to identify the most \textit{informative} portions of notes to inform predictive models\cite{Murdoch2013-wx}. 

Current EHR system designs have focused on the administrative side of care delivery forcing clinical users to spend more of their time performing data entry \cite{Tai-Seale2017-hj,Ratwani2019-iy,Holmgren2021-eu}. Systems do not allow users to refer to, append, or amend prior notes whilst keeping the original document as recorded \cite{Bowman2013-xx}. To overcome this limitation free text is often copied from prior notes, duplicating data that could otherwise be referenced \cite{ODonnell2009-ye}.

This work aims to highlight and quantify an often acknowledged but neglected area of study - the scale of redundancy in EHR text. As redundancy is so prevalent in clinical text the research community must do more to understand where and why this redundancy exists in an effort to minimise and mitigate its effects, allowing for further progress in the diverse use cases of clinical text as previously discussed.

Understanding where the most meaningful data is within a record will enable researchers to better understand where time should be spent preparing data, as well as potentially informing EHR system designers where changes can be made to improve data entry design or other data redundancy reduction mechanisms for future implementations. 

We present two approaches to measure redundancy in clinical texts:
\begin{itemize}[leftmargin=*]
    \item \textbf{Information-theoretic redundancy:} 
    We show language models trained and tested on public and private clinical texts consistently show higher levels of redundancy in comparison to open-domain text as demonstrated by information-theory measures of perplexity and cross-entropy\cite{Shannon1951-vu}.
    \item\textbf{Syntactic and semantic redundancy of successive note pairs:} we show average token level redundancy across various clinical note types, through calculation of summarisation metrics of temporally successive note pairs. This measure assumes that successive notes from the same admission and of the same type are `summaries' of former notes within the same clinical admission. We discuss the implications of recall and precision of these metrics and perform a manual analysis of randomly selected notes.
\end{itemize}

\section{Background}

\subsection{Prior Work}
Despite information redundancy in clinical text being widely reported, work to develop  methods or measures of redundancy and applying these to clinical text have been limited. Early work investigated lexical matching to measure redundancy \cite{Wrenn2010-ci}, presenting a modified Levenshtein edit-distance based algorithm that aligned and measured redundancy of 100 randomly selected admissions \cite{Wrenn2010-ci} reporting an average 78\% and 54\% redundancy for sign-out and progress notes respectively. Further work applied lexical normalisation, stop word removal followed by a sliding window alignment algorithm over multiple sentences \cite{Zhang2011-go}, showing a 82\% correlation with human annotated expert judgements of redundancy for randomly selected sentences in outpatient notes. 

Assessing the semantic similarity of documents provides a more robust method to detect redundancy, as lexical and syntactic variations that may arise when a prior note is summarised or copy/pasted then edited can still be marked redundant. Prior work has used statistical modelling techniques to recognise new relevant information for various note types \cite{Zhang2014-nc,Zhang2017-ya}. 

Automated summarisation systems perform a similar process to redundancy identification. Intuitively, an effective summary will identify the most `important' sections of a document, highlighting the informative, relevant parts of a document whilst ignoring the redundant sections \cite{Peyrard2019-xt}. An extractive summary of text can be seen as an inverse ranking of redundancy, selecting the least redundant sections of a source text, and an abstractive summarisation performs the same ranking followed by a natural language generation step \cite{Moen2016-js}. Outside of the clinical domain, there is strong interest in models for open-domain free text summarisation \cite{See2017-fx,Raffel2020-tj,Lewis2020-gj,Zhang2020-kv}. Many of these methods use deep neural network based methods to learn representations that capture lexical, syntactic and semantic meaning of texts to produce coherent and informative summaries. Most methods are \emph{knowledge-free}, having no reliance on external modelled knowledge graphs or databases and learn to write summaries only from input text    and the associated reference summary.

The clinical domain is uniquely rich with modelled knowledge graphs such as the UMLS \cite{Bodenreider2004-ci} and SNOMED-CT \cite{Stearns2001-yt}. Applying Named Entity Recognition and Linking systems such as cTakes \cite{Savova2010-df}, MetaMap \cite{Aronson2001-kr} or MedCAT \cite{Kraljevic2021-ln} over EHRs and aggregating extracted concepts over groups of documents per admission could determine documents with equivalent extracted concepts as redundant. However, solving such an NER+L task is an ongoing research problem due to the scale of modelled knowledge (i.e. hundreds of thousands of possible concepts) and the variability of clinical text \cite{Wu2015-zo,Wu2017-yk}.

Recently, corpora of synthetic \cite{Rastegar-Mojarad2018-yg} and manually annotated\cite{Wang2020-vl} semantic similarity sentence pairs have been used in shared tasks to promote further research and system development in this area \cite{Wang2020-so}. Deep neural models such as BERT \cite{Devlin2019-ny} and S-BERT \cite{Reimers2019-ip} achieved high scores from multiple challenge submissions achieving 0.88 correlation in ranking sentences with a similarity scale of 0-5.

To our knowledge there is no prior work that estimates information theoretic content of clinical text and compares such estimates to open-domain text. Prior work has estimated redundancy using sequence alignment algorithms for estimating token-level redundancy, largely not considering semantic redundancy, i.e. the tokens differ across texts but the meaning is equivalent, or they have considered sentence to sentence semantic similarity, training models to predict similarity between sentences.

\subsection{Measuring Redundancy of Text through Informativeness}
The following sections provide the information theoretic basis for empirically estimating redundancy of clinical text. We initially introduce relevant notation and information theory concepts, then describe how language modelling can be used to estimate redundancy. 

Given a language $L$ with a vocabulary $V$ comprised of the number of $n$ symbols $w_1 \ldots w_n \in V$ where $w_i$ is a character, word or \emph{word piece} produced by some tokeniser function $Z$ over text $t$, $Z(t)$ provides some sequence of $w$ symbols. Given that $P$ is a probability distribution over all symbols in $V$ we can define the average \emph{information} conveyed by a language $L$ via Shannon's Entropy \cite{Shannon1997-wu}. $H(P)$ is defined as:

\begin{equation} \label{eq:entropy}
    H(P) = E[I_2(P)] = -\sum^{n}_{i=1}p(w_i) \log_2 p(w_i)
\end{equation}

Entropy is the negative sum of proportional $\log_2$ probabilities of each symbol $w_i$ with information units represented as bits (i.e. $log_2$). Intuitively, entropy provides the average number of bits used to convey a symbol from set $V$ for the most efficient coding of $L$. A maximum bound for the entropy of $L$ is the uniform distribution for $P$ over all symbols in $V$. Given Equation \ref{eq:entropy} this provides:

\begin{equation}\label{eq:uniform_entropy}
    \begin{split}
        H(P) &= \sum^{n}_{i=1} p(w_i) \log_2 p(w_i)\\
        &= \frac{1}{n}\sum^{n}_{i=1}\log_2n = \frac{1}{n}n\log_2n \\
        &= \log_2n \\
    \end{split}
\end{equation}

A theoretical lower bound of $H(P) \approx 1$ is if the probability of a single symbol $W$ is ${P(W = w_i) \approx 1}$ as the probability mass is focused on $w_i$, i.e. $L$ effectively only has 1 symbol. Equation \ref{eq:entropy} holds in the limit of all possible texts that can be produced for $L$. As we cannot produce all possible texts from $L$ we empirically estimate $H(P)$ with a distribution $Q$ over the same vocabulary $V$ for some, usually large, defined set of texts from $L$. The cross entropy between distributions $P$ and $Q$ is:

\begin{equation}\label{eq:dkl}
    H(P, Q) = H(P) + D_{KL}\left(P \middle\| Q\right)
\end{equation}
where $D_{KL}\left(P \middle\| Q\right)$ is the Kullback-Leibler(KL) divergence or relative entropy of Q from P. These are the extra bits needed to encode symbols from distribution P through the use of the optimal encoding scheme found through the distribution Q.

\subsection{Causal Language Modelling}
Causal Language modelling (LM) is the task to predict the next symbol conditioned on previous symbols. Given a defined set texts from $L$ fitting such a model minimises the $D_{KL}\left(P \middle\| Q\right)$ term of Equation \ref{eq:dkl} therefore providing an estimate of entropy for $L$. A language model estimates the joint probability of a sentence by conditioning the current symbol $w_i$ on all previous $w_1 \ldots w_{i-1}$:

\begin{equation}
    P(w_1, ..., w_i) = p(w_1) ... p(w_{i}| w_{1}, ..., w_{i-1})
\end{equation} 

\subsection{Perplexity and Cross-Entropy to Compare Redundancy Across Texts}
 Perplexity (PPL) is the `surprise' a language model finds having encountered $w_{n}$ given $w_{1}, \ldots w_{n-1}$, and is the $2^{H(P, Q)}$ of entropy\cite{Jurafsky2009-yb}. Language models are often evaluated using PPL where the lower the score the better the model generalises to unseen texts from language $L$. Given a language model trained on general purpose text $L_{en}$, and another language model with the same available vocabulary $V$ trained on clinical text $L_{clinic}$ then comparing PPL / i.e. cross-entropy by taking $log_2$(PPL), provides a reflection of the level of \textit{information} and therefore redundancy present in texts across the two languages. 
 
 It is however important to highlight that this information theoretic measure of redundancy, i.e. estimating the efficiency of encoding of a given a language given the same language model, does not capture a human level measure of informativeness as clinical texts are subject to a context in which they are written. For example, clinical text progress reports have represent a time series of clinical information and therefore repetitions in text could indicate a continuation or confirmation of prior clinical information and may not necessarily be redundant.

\subsection{Re-purposing Summarisation Evaluation Metrics for Sequential Note Sequences}
The primary purpose of clinical narratives are to document new clinical information. However, EHR data entry often is often poorly designed \cite{Bloom2021-ir} or users lack sufficient training, time or incentives for clean data entry. This results in frequent use of the copy-paste function with prior data copied into the current note with additions and amendments for the new clinical information\cite{Hirschtick2006-gw,ODonnell2009-ye,Venkateshaiah2010-uf}. Therefore, our second set of experiments frame a set of clinical notes of the same type for a given admission as successive summaries of one another and seeks to measure the prevalence of copy-pasted notes from successive note pairs.

We apply n-gram and semantic embedding summarisation metrics to successive pairs of clinical notes. In this context `recall' captures the proportion of the previous note that is contained in the current note, whereas `precision' is more ambiguous as successive notes with high precision and high recall indicate a note is redundant (i.e. the content is equivalent), whereas high recall, low precision indicates a summary of the previous note with additional new information. Low recall and low precision indicates a successive note does not summarise prior events at all, we expect this to be the case for procedure and investigative notes such as radiology reports as these events are often standalone, even if they take place during the same admission. There are no clear aims for high precision / recall such as the case for comparing predictive model performance. 

\section{Methods}\label{sec:methods}

\subsection{Datasets}\label{sec:datasets}
Descriptive statistics for datasets and splits are provided in Table \ref{tab:desc_stats}. We consider two clinical datasets in our analysis, we take a `stroke' specific subset to compare results to our other clinical dataset:
\begin{itemize}[leftmargin=*]
    \item MIMIC-III: \cite{Johnson2016-mq} A large, freely-available US based ICU dataset collected between 2001-2012 containing 53,423 distinct admissions. We consider MIMIC-FULL ($\sim$1.17M documents) that contains all free text notes for primary coded conditions that appeared at least 20 times ($\sim$41k admissions), and MIMIC-Stroke (337 admissions) with a primary diagnosis of ICD10 code:I63.*. 
    \item KCH: clinical records for patients diagnosed with Cerebral infarction (ICD10 code:I63.*) from the King's College Hospital (KCH) NHS Foundation Trust, London, UK, EHR. This includes 9,892 distinct admissions and $\sim$26K  documents. We extract data via the internal CogStack \cite{Jackson2018-km} system, an Elasticsearch based ingestion and harmonization pipeline for EHR data. This patient cohort is driven by permitted ethical approval and our ability to compare to a similar patient cohort in MIMIC-Stroke.
\end{itemize}

Our two open domain English language datasets are available via the HuggingFace Datasets\footnote{https://huggingface.co/docs/datasets/master/} library, and are used to demonstrate the entropy / PPL of non-clinical open-domain datasets. We use:
\begin{itemize}[leftmargin=*]
    \item OpenWebText \cite{Gokaslan2019-km}: a recreated openly available version of the original data used to train GPT-2. There is no defined 'test' split so we randomly sample 5000 texts. It is worth noting our base pre-trained language model (GPT-2 \cite{Radford2019-hk}) has likely seen some if not all of the samples in this random sample during pre-training. Vocabulary size is 48,105.
    \item WikiText2 \cite{Merity2017-ra}: the test data split of WikiText2, a corpus of 4358 Wikipedia articles often used to assess language models. This data is unseen by all LMs and is used to assess open-domain text language modelling performance.
\end{itemize}

\setlength\tabcolsep{1pt} 
\begin{table}[]
    \centering
    \begin{tabular}{ccccc}\toprule
        \textbf{Dataset} & \textbf{\# Docs} & \textbf{\shortstack{Avg.\\ Length}} & \textbf{\shortstack{\# Note\\ Types}} & \textbf{\shortstack{Test Set\\Vocab Size}} \\
        \midrule
        M-III & 1,172,433 & 2,201 & 3,127 & 31,017 \\ 
        M-III (S) & 8,213 & 2,232 & 241 &  12,167 \\   
        KCH & 26,348  & 5,217 & 1310 & 27,722 \\
        WebText & 5000 & n/a & n/a & 48,105 \\
        WikiText2 & 4358 & 579 & n/a & 19,037 \\
        \bottomrule
    \end{tabular}
    \caption{Descriptive statistics for clinical and open domain datasets. Average document length is in characters and a single note type for MIMIC-III is the combined \emph{category} and \emph{description} fields. KCH uses a single field for note type. M-III is the MIMIC-III 'full' dataset and (S) is the stroke (I63.*) primary diagnosis subset. WebText \& WikiText-2 do not have \# 'Note Types' and WebText is only available as sentences only.}
    \label{tab:desc_stats}
\end{table}

\subsection{Experimental Setup}

\subsection{Data Preparation}
To exclude very rare conditions or cases that may not represent typical clinical language found in EHRs we extract all MIMIC-III notes and filter the admissions that have a primary diagnosis that appeared $\ge$ 20 times in the dataset. We decided upon this threshold after initial small-scale experimentation. We do not clean the notes from MIMIC-III or KCH in any way, although the MIMIC-III notes have already undergone a de-identification process to remove sensitive information such as dates and names.

\subsubsection{Pre-trained Language Models}
We estimate the entropy of clinical language using GPT-2 \cite{Radford2019-hk} a previous state-of-the-art auto-regressive causal language model, based upon the Transformer \cite{Vaswani2017-db} architecture that has been pre-trained with the `WebText' corpus, $\sim$40Gb of text data collected from the Web. Model / tokenizer weights, configurations and model implementations are via the HuggingFace `transformers' \cite{Wolf2020-dm} library. We use the base GPT-2 model with 124M parameters, 12 Transformer block layers with model dimensionality of 768, and vocabulary size 50,257. 

\subsubsection{Language Model Fine Tuning and PPL Calculations}
We fine-tune GPT-2 in a self-supervised manner, i.e. after tokenizing the clinical text we feed each token sequentially into the model, conditioning on previous symbols, we produce the distribution over $V$ via the forward pass of the model, compute the loss and back-propagate the error gradient back through the model to update parameters. Code for tokenizing, training, validating and testing the fine-tuned model for the openly available datasets are made available\footnote{https://github.com/tomolopolis/clinical\_sum}. We calculate perplexity by concatenating all test set texts and applying a strided sliding window half the size of the model dimension (384) to condition the model and make a token prediction. This method ignores inconsistent sentence breaks, a common problem in EHR text. Importantly, this produces results inline with original GPT-2 \cite{Radford2019-hk} work, allowing us to focus on the impact the datasets have on PPL calculations. 

\subsubsection{Internote Type Summary Evaluation}\label{sec:sum_metrics_algos}
\begin{figure*}
    \centering
    \includegraphics[width=\linewidth]{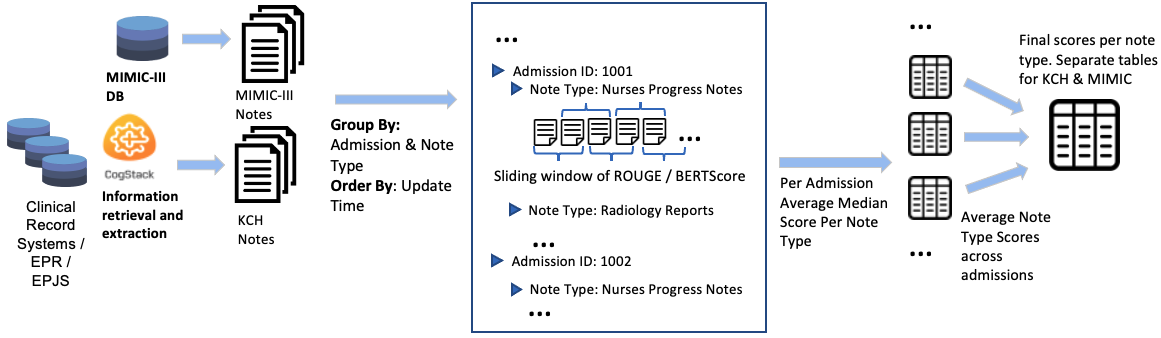}
    \caption{Internote type summarisation evaluation process.}
    \label{fig:token_redund_summ_process}
\end{figure*}

Our second method of estimating levels of redundancy in clincal text applies summarisation evaluation metrics to ordered note pairs as demonstrated in Figure \ref{fig:token_redund_summ_process}. We firstly group each admission's note types and order by update time. We apply a sliding window of pairwise evaluations over each note sequence then average over the sequence and admissions. Our output is a table for MIMIC and KCH with the average token level summarisation score per note type. This method measures the level of redundancy between successive clinical notes within the same admission of the same type.

We use a Gestalt Pattern matching algorithm\cite{Black2004-jl} as a baseline that computes the ratio of matching sub-sequences of `tokens', (i.e. white-space separated words) between each successive note. We then report precision/recall for ROUGE \cite{Lin2004-my} another lexical/syntactic token metric and BERTScore \cite{Zhang2020-lf} a recent deep-learning model based metric that embeds texts using pre-trained semantic vector space, cosine similarity between the embedded texts produces a similarity score between them. BERTScore was shown to correlate higher with human level judgements of generated summary quality than token based metrics such as ROUGE, somewhat addressing the documented failings of ROUGE \cite{Schluter2017-ov}. Our clinical texts are longer than the maximum dimension supported by the default and highest performing model configured with BERTScore. Therefore, We use the xlnet-base-cased \cite{Yang2019-ye} embeddings due to increased maximum permitted input length. Our scores are normalised to the model baseline to produce an improved uniformity in similarity scores as discussed in the original work \cite{Zhang2020-lf}.

\section{Results}
We present results for both clinical datasets presented in Section \ref{sec:methods} and open datasets originally used to train/test LMs. 

\subsection{Estimating Entropy of Clinical Text}
Table \ref{tab:ppl_results} reports PPL scores across datasets used to pre-train and further fine-tune GPT-2 models. We report our test set results for the pre-trained GPT-2 and the model fine-tuned to clinical datasets presented in Section \ref{sec:datasets}. `Test' values for each dataset provide empirical estimates of entropy for languages $L_{en}$ i.e. OpenWebText, and $L_{clinic}$, i.e. MIMIC (Stroke / Full) and KCH. 

\setlength\tabcolsep{6pt}
\begin{table}[]
    \centering
    \begin{tabular}{ccc||c}\toprule
        \textbf{Dataset} & \textbf{Val} & \textbf{Test} & \textbf{WikiText2} \\\midrule
        OpenWebText & - & 29.57  & 35.56  \\
        MIMIC (Stroke) & 6.14 & 5.38 & 144.4 \\
        MIMIC (Full) & 3.12 & 3.15 & 204.9  \\
        KCH    &  8.78  & 9.58  & 74.51 \\\bottomrule
    \end{tabular}
    \caption{Perplexity scores for GPT-2 trained on (Open)WebText (i.e. the model is not trained in this work at all), further training on the MIMIC (Stroke), KCH, and MIMIC (Full) datasets. WikiText2 test split results are also provided for an unseen test set of open-domain text for all models.}
    \label{tab:ppl_results}
\end{table}

We show LM performance on validation and test sets, observing that test set PPLs are largely consistent with validation set scores indicating the models are not over-fitting to idiosyncrasies only present in the validation set. We are potentially underfitting the data as we did not especially experiment with techniques such early stopping, learning rate optimisation and architecture optimisation. As the model performance is not the valuable contribution of this work we only used a small number of fixed epochs (i.e. 8) with a scheduled weight decay within the AdamW \cite{Loshchilov2019-zn} optimizer (i.e. 0.01).

Our results demonstrate the PPL of clinical texts to be smaller than open domain text. Using Equations. \ref{eq:uniform_entropy}, \ref{eq:dkl} and computing $log_2(PPL)$ we estimate the \emph{information content} of our open-domain text language $L_{en} = 5.16$ and our clinical language $L_{clinic} = 1.66-3.26$. This suggest that clinical text is $\sim1.5$x to $\sim3$x less efficient in encoding \emph{information} than regular open domain text. It is important however to note this \textit{efficiency} is with the respect to the definition of an optimal encoding of a language $L$. Predictability of texts within $L_{clinic}$ does not necessarily measure the informativeness from a human perspective in comparison to $L_{en}$.

We further test our models on WikiText-2 dataset to observe open-domain performance after clinical text training. We find that once GPT-2 is further trained with clinical text it loses the ability to accurately model open-domain text resulting in large PPLs. This is seen to a greater extent in MIMIC (Full) compared to MIMIC (Stroke) / KCH, which is likely due to the MIMIC (Full) model having seen the highest volume of clinical text. 

\subsubsection{Perplexity Across Clinical Datasets}
We compare our models trained and tested on  available alternative clinical datasets as shown in Table \ref{tab:cross_clinic_ds_results}. As our MIMIC (Stroke) / KCH trained models share the common stroke diagnosis we would expect clinical language and the description of symptoms, findings, clinical events, procedures to be similar. Our KCH trained and MIMIC (Stroke) tested model performs modestly, i.e. PPL is still 6-13 points less than open domain PPLs, whereas the MIMIC trained and KCH tested model performs poorly. Surprisingly, the similarity in disorder seems to offer little or no benefit, as KCH trained and testing on both MIMIC test sets produces similar PPLs. MIMIC trained and KCH tested also performs better with \emph{Full} compared with \emph{Stroke}. We believe the poor performance with MIMIC trained models is due to heterogeneity of the KCH dataset, including out patient notes, patient letters, procedure reports etc. whereas MIMIC only contains inpatient ICU notes albeit notes from across specialisms such as physician, nursing, radiology, etc. 

\setlength\tabcolsep{1.5pt} 
\begin{table}[]
    \centering
    \begin{tabular}{ccc}\toprule
        Training & Test  & PPL \\\midrule
        KCH & MIMIC (Stroke) & 23.05 \\
        KCH & MIMIC (Full) & 23.98 \\
        MIMIC (Stroke) & KCH & 119.66 \\ 
        MIMIC (Full) & KCH & 94.19 \\\bottomrule
    \end{tabular}
    \caption{GPT-2 trained and tested across our clinical datasets.}
    \label{tab:cross_clinic_ds_results}
\end{table}

\subsection{Token Level Redundancy}
\begin{figure*}[!ht]
\begin{subfigure}{.5\textwidth}
  \centering
  \includegraphics[width=.98\linewidth]{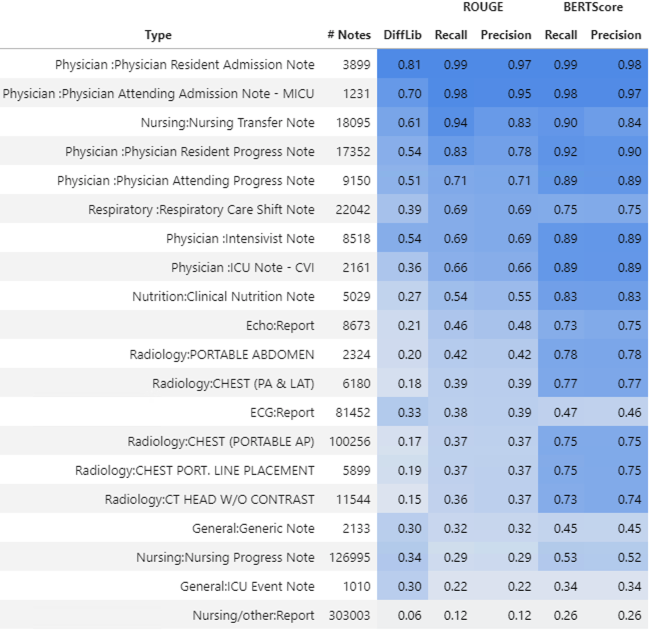}
  \caption{MIMIC-III (Full) Summarisation metrics by type}
\end{subfigure}%
\begin{subfigure}{.5\textwidth}
  \centering
  \includegraphics[width=.98\linewidth]{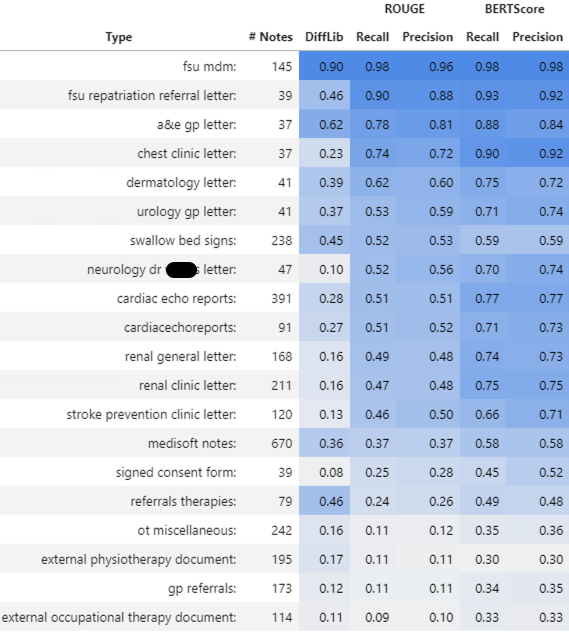}
  \caption{KCH Summarisation metrics by type} 
\end{subfigure}
\caption{Summarisation metrics calculated over a sliding window of \emph{generated} and \emph{reference} summaries for admission texts grouped by admission then by note type and ordered by time. We only show the first 20 note types of each dataset ordered by ROUGE score.}
\label{fig:summarisation metrics}
\end{figure*}

Figure \ref{fig:summarisation metrics} shows our results computing summarisation metrics described in Section \ref{sec:sum_metrics_algos} for the MIMIC (Full) and KCH datasets. Broadly, our baseline (difflib), ROUGE and BERTScore metrics display similar trends, as seen by coloured gradients consistently decreasing across all metrics for similar types of documents. There are some exceptions in the MIMIC dataset such as \emph{Respiratory: Respiratory Care Shift Note} where our baseline method reports a lower similarity ratio as compared to the summarisation metrics.

We report the micro-averaged median scores for each note type to reduce skew from extremes of either side of the distribution of scores. Recall and precision for ROUGE and BERTScore at each note type are largely equivalent, indicating each note type has on average proportionally equivalent amounts of redundant, i.e. duplicated text, from previous notes (the recall score), and `new' text (the precision score. We observe that this varies substantially according to note type with almost no redundant text with some types, i.e. \emph{Nursing/other:Report} and in contrast the majority of text being redundant, i.e. \emph{Physician:Physician Resident Admission Note}.

Table \ref{tab:final_redun_across_metrics} shows a final average across each metric weighted by total number of tokens within each document and type. Interestingly, recall and precision are equivalent for ROUGE and BERTScore. Intuitively, this indicates that successive notes often have a `core' section which is static throughout an admission and updates are provided by editing certain sections only. This reflects a typical workflow for providing status updates on patient condition or progress.

\setlength\tabcolsep{2.5pt} 
\begin{table}[]
    \centering
    \begin{tabular}{c c c c c c}\toprule
        \textbf{Dataset} & \textbf{DiffLib} & \multicolumn{2}{c}{\textbf{ROUGE}} & \multicolumn{2}{c}{\textbf{BERTScore}} \\
                &         & \textbf{Rec} & \textbf{Prec} & \textbf{Rec} & \textbf{Prec}\\\midrule
        MIMIC & 0.26 & 0.43 & 0.42 & 0.58 & 0.58 \\
        KCH & 0.32 & 0.49 & 0.49 & 0.65 & 0.65\\\bottomrule
    \end{tabular}
    \caption{Weighted average by token length of sequential token level redundancy. Rec = Recall, Prec = Precision.}
    \label{tab:final_redun_across_metrics}
\end{table}

\subsection{Manual Analysis}
We perform a manual analysis of 70 randomly selected note pairs, (35 each from MIMIC-III and KCH). We group, order and split the notes as shown in Figure \ref{fig:token_redund_summ_process} and visually highlight the token level differences between successive pairs to assist with determining similarity / differences. We use a Likert scale of 1-5 to rate redundancy between note pairs and compute a correlation with F1 score. Table \ref{tab:manual_analysis} shows that ROUGE scores correlate better with our human annotated measure  of redundancy than BERTScore.

\setlength\tabcolsep{6pt} 
\begin{table}[]
    \centering
    \begin{tabular}{c c c}\toprule
        \textbf{Dataset} & \textbf{ROUGE} & \textbf{BERTScore} \\\midrule
        KCH & 0.83 & 0.77 \\
        MIMIC & 0.77 & 0.63 \\
        \bottomrule
    \end{tabular}
    \caption{F1 score correlation of redundancy of ROUGE and BERTScore with manual annotations on a 1-5 Likert scale of a random sample of note pairs.}
    \label{tab:manual_analysis}
\end{table}

\section{Discussion}

\subsection{Language Modelling for Clinical Text}
Our PPL scores suggest that clinical text is $\sim1.5$x to $\sim3$x less efficient in encoding information than regular open domain text, or $\sim1.5$x to $\sim3$x more text is used to communicate the same volume of \emph{information} in comparison to open domain text.

To our knowledge this is the first work to estimate in information theoretic terms the entropy of clinical language $L_{clinic}$ and compare against open-domain language $L_{en}$. These estimates are dependent upon the text and models used, but we believe they are representative as both datasets are large, from varied geographies, hospital sites, specialisms and patient types (outpatient vs inpatient). Our $L_{en}$ corpora are built from curated texts (i.e. Wikipedia and positive karma Reddit posts) that cover a wide array of topics. However, our results may be highly dependent upon these text sources. Future work could compare other easily available datasets such as news or academic papers to provide further clarity on our findings.

Language modelling performance is dependent upon the size of vocabulary of the model and the test set. Model vocab size is static as the same model (GPT-2) and tokenizer configurations are used throughout all experiments. Despite the narrower focus of clinical text, the vocabulary sizes in Table \ref{tab:desc_stats} indicate MIMIC-III (Full) and KCH are in fact larger than the WikiText2 corpus although we observe substantially lower PPLs for clinical text. This suggests clinical text is overall less \textit{informative} and therefore more redundant when compared to open-domain corpora. However, this interpretation must be further clarified, as EHRs are written with a clear task in mind to communicate health status, and record clinical events. This is in contrast to open-domain text that has a far wider array of possible tasks for the text.

We compute PPL scores inline with the original GPT-2 authors \cite{Radford2019-hk}, as this work is an assessment of the data rather than the specific model. A reduced sliding window stride length during PPL calculation would decrease scores further, although relative difference would remain similar. However, we acknowledge that our results are dependent on model architecture, i.e. GPT-2 has higher performing model variants `GPT-2(large)' even newer variants, `GPT-3', with an even larger parameter space \cite{Brown2020-ft} We propose our results show the trend that clinical domain text is redundant by some multiple compared to open-domain text.

The drop in open-domain text performance after clinical text fine-tuning suggests the model is incapable of modelling clinical and open-domain text simultaneously. The difference in lexicon and syntax forces the model to minimise a loss landscape substantially different from that found in open-domain text. Further work, could experiment with larger models or with a training process that jointly attempts to model open-domain and clinical text, in an effort to maintain high performance on both. Multiple works \cite{Radford2019-hk,Raffel2020-tj} have already highlighted the effect and importance of data quality, pre-processing and training configuration in LM training. 

\subsection{Sequential Inter-Note Type Redundancy} 
We used BERTScore configured with \emph{xlnet-base-cased}, due to the size of input texts. The \emph{xlnet-base-cased} embeddings in the BERTScore framework report worse correlation with human annotations of summarisation quality than the default settings that otherwise do not support long input texts. Our manual evaluation of notes against the computed scores indicate ROUGE more accurately captures redundancy than the current BERTScore configuration. During the manual review we noticed BERTScore often scored notes highly that had small token-level differences. As BERTScore projects note pairs into a learnt semantic vector space it is difficult to compare scores with the n-gram based ROUGE. One explanation is that note pairs are likely by the same clinician, are the same clinical specialism and about the same patient and therefore score highly, although n-gram differences are larger. A model such as ClinicalXLNET \cite{Huang2020-kl} would likely assist in capturing differences in clinical language thereby producing more appropriate embeddings compared to the open domain variety currently used. We leave this experiment to future work.

This work only considers sequences of notes labelled as the same type. Analysis of intra-note type redundancy, where notes of one type refer to clinical events documented in other note types is another potential avenue of future work. Future work could also order note sequences by clinician, or compare only first and last note for example.

Overall, the interpretation of recall and precision of the summarisation metrics and their relationship to redundant text is nuanced. For example, repeated mentions of an acute condition may simply indicate the continued presence of a condition or symptom, and may not be redundant text after all. These measures do not account for the time series nature of clinical information present in the record. Future work could investigate information extraction, normalisation and linking methods that leverage clinical knowledge bases such as cTakes\cite{Savova2010-df}, MetaMap\cite{Aronson2001-kr} or MedCAT\cite{Kraljevic2021-ln}. Extracted concepts could then be compared across notes whilst being grounded in clinical knowledge. This would allow for redundant clinical events to be identified alongside how they present in the text.




\section{Conclusions}
We have presented two empirical approaches for an often acknowledged \cite{Murdoch2013-wx} but neglected area of clinical natural language processing research, to measure redundancy in clinical text. We have trained large language models on multiple clinical datasets resulting in perplexity and therefore cross-entropy estimates for a clinical language $L_{clinic}$. We observe a $\sim1.5$x to $\sim3$x reduction in entropy when comparing the same model trained on open domain text. Our approach shows the token level redundancy between different note types with the usage of automated summarisation evaluation metrics. We observe variable scores across different types with some results indicating clinical notes can be 97-98\% redundant (i.e. the text is largely duplicated across documents \emph{MIMIC: Physician Resident Admission Note}), or only 0.12\% redundant (\emph{MIMIC: Nursing/other:Report}). 

Overall, our results support prior work suggesting clinical text contains redundant text  \cite{Murdoch2013-wx,Wrenn2010-ci,Zhang2011-go}. In information theory terms we show that clinical text is less \emph{efficient} than open domain text meaning on average more text is required to express the same volume of \emph{information} in comparison to general purpose texts. However, this \textit{efficiency} measure does not take into account the context in which EHR records are written, that is a time series of clinical events, where repetition may not necessarily be redundant but indicative of an ongoing condition or clinical event.

With more stressors on our healthcare system than ever before  \cite{Mesa_Vieira2020-bn} and despite increasing investment \cite{Jakovljevic2020-ox} we continue to see increased clinician burn-out \cite{Montgomery2019-nd}. A contributing factor is the often enforced usage of EHR systems, increasing doctor-computer time \cite{Kroth2018-uy}, forcing clinicians to overcome poor usability of systems \cite{Bloom2021-ir}. Improving EHR entry to allow easy updating, cross referencing and versioning of notes could alleviate an extra burden on clinical staff. To this aim we would urge EHR providers to adapt their systems to improve data entry and maintenance, potentially considering features similar to source code management version control allowing for a \emph{living} document to improve data quality, minimise redundancy and errors that are propagated through the usage of copy/paste. We acknowledge this would however require substantial non-trivial changes to systems and user workflow \cite{Lyons2011-kv,Schmucker2009-vi}. Until EHR providers address these shortcomings researchers will have to rely on ad-hoc pre-processing logic to clean datasets before carrying out analysis.

\section*{Data Availability Statement}
Open-domain text (OpenWebText and WikiText2) data is openly available as described in Section \ref{sec:datasets}. MIMIC-III\cite{Johnson2016-mq} is freely available but users must obtain permission and a license from dataset owners. KCH data is a highly sensitive dataset and is not easily available. Interested researchers are encouraged to discuss potential projects with the authors to discuss how data access can be granted. 

\section*{Acknowledgements}
RD's work is supported by 1.National Institute for Health Research (NIHR) Biomedical Research Centre at South London and Maudsley NHS Foundation Trust and King’s College London. 2. Health Data Research UK, which is funded by the UK Medical Research Council, Engineering and Physical Sciences Research Council, Economic and Social Research Council, Department of Health and Social Care (England), Chief Scientist Office of the Scottish Government Health and Social Care Directorates, Health and Social Care Research and Development Division (Welsh Government), Public Health Agency (Northern Ireland), British Heart Foundation and Wellcome Trust. 3. The National Institute for Health Research University College London Hospitals Biomedical Research Centre. This paper represents independent research part funded by the National Institute for Health Research (NIHR) Biomedical Research Centre at South London and Maudsley NHS Foundation Trust and King’s College London. The views expressed are those of the author(s) and not necessarily those of the NHS, MRC, NIHR or the Department of Health and Social Care.

\bibliography{custom}
\bibliographystyle{acl_natbib}

\appendix



\end{document}